\documentclass{article} 
\usepackage{iclr2026_malgai,times}

\usepackage{amsmath,amsfonts,bm}

\def\eqref#1{equation~\ref{#1}}

\def\1{\bm{1}}

\DeclareMathAlphabet{\mathsfit}{\encodingdefault}{\sfdefault}{m}{sl}
\SetMathAlphabet{\mathsfit}{bold}{\encodingdefault}{\sfdefault}{bx}{n}

\usepackage{hyperref}
\usepackage{url}
\usepackage{xurl}
\usepackage{graphicx}
\usepackage{booktabs}
\usepackage{microtype}
\usepackage{enumitem}
\usepackage{amsmath,amssymb}
\usepackage{mathtools}
\usepackage{algorithm}
\usepackage{algpseudocode}
\usepackage{xcolor}
\usepackage{needspace}
\usepackage{fvextra} 
\usepackage{placeins}
\usepackage{array}
\usepackage{authblk}

\makeatletter

\renewcommand\thefootnote{\fnsymbol{footnote}}
\makeatother

\newcommand{\NOMI}{\textsc{MAGIC}}
\newcommand{\GEPA}{\textsc{GEPA}}
\newcommand{\MAMUT}{\textsc{MAMuT}}

\newcommand{\trace}{\tau}
\newcommand{\TraceSet}{\mathcal{T}}

\newcommand{\Judge}{\mathcal{J}}
\newcommand{\Rubric}{\mathcal{R}}

\newcommand{\Prompts}{\mathcal{P}}
\newcommand{\Score}{\mathsf{S}}

\title{Build, Judge, Optimize: A Blueprint for Continuous Improvement of Multi-Agent Consumer Assistants}
\newcommand{\equalcontrib}{\textsuperscript{*}}

\author[1]{Alejandro Breen Herrera\equalcontrib}
\author[1]{Aayush Sheth\equalcontrib}
\author[2]{Steven G.~Xu\equalcontrib}
\author[2]{Zhucheng Zhan}
\author[1]{Charles Wright}
\author[1]{Marcus Yearwood}
\author[2]{Hongtai Wei}
\author[2]{Sudeep Das}
\author[2]{Danny Nightingale}
\author[2]{Meg Watson}
\author[2]{Charles Pollnow V}

\affil[1]{Metis}
\affil[2]{DoorDash}
\iclrfinalcopy 
\begin{document}
\maketitle
\begingroup
\renewcommand\thefootnote{\fnsymbol{footnote}}
\footnotetext[1]{%
  \begin{minipage}[t]{\linewidth}\raggedright
    Equal contributions. Correspondence to
    \{alejandro, aayush\}@withmetis.ai,
    steven.xu@doordash.com
  \end{minipage}%
}

\begin{abstract}
    Conversational shopping assistants (CSAs) represent a compelling application of agentic AI, but moving from prototype to production reveals two underexplored challenges: how to evaluate multi-turn interactions and how to optimize tightly coupled multi-agent systems. Grocery shopping further amplifies these difficulties, as user requests are often underspecified, highly preference-sensitive, and constrained by factors such as budget and inventory. In this paper, we present a practical blueprint for evaluating and optimizing conversational shopping assistants, illustrated through a production-scale AI grocery assistant. We introduce a multi-faceted evaluation rubric that decomposes end-to-end shopping quality into structured dimensions and develop a calibrated LLM-as-judge pipeline aligned with human annotations. Building on this evaluation foundation, we investigate two complementary prompt-optimization strategies based on a SOTA prompt-optimizer called GEPA~\citep{shao2025_gepa}: (1) \emph{Sub-agent GEPA}, which optimizes individual agent nodes against localized rubrics, and (2) \emph{\MAMUT} (\textbf{M}ulti-\textbf{A}gent \textbf{Mu}lti-\textbf{T}urn) GEPA~\citep{mamuth-gepa2026}, a novel system-level approach that jointly optimizes prompts across agents using multi-turn simulation and trajectory-level scoring. We release rubric templates and evaluation design guidance to support practitioners building production CSAs.
\end{abstract}

\section{Introduction}
Agentic AI systems are increasingly deployed in applications requiring sustained interaction, tool use, and autonomous reasoning. Conversational shopping assistants (CSAs) exemplify this shift, transforming e-commerce from keyword-based search into collaborative, dialogue-driven experiences~\citep{griddynamics2025_agentic_commerce}. This trend is reflected in emerging systems such as Amazon’s Rufus~\citep{amazon_rufus_2024} and Google Shopping’s AI mode~\citep{google_shopping_ai_mode_2025}. Rather than manually filtering products, users express high-level goals while the assistant interprets intent and reasons across multiple turns. 

Traditional retrieval and ranking metrics are insufficient for CSAs, where quality is multi-dimensional and must be assessed across multi-turn interaction trajectories. Optimization is equally challenging: improving individual sub-agents does not reliably translate to better end-to-end outcomes due to delayed effects and cross-agent coupling. Despite growing interest~\citep{zhu2025multiagentbench, sun2025llmagentmeetsagentic}, systematic methods for evaluating and optimizing production-scale multi-agent CSAs remain limited.

In this paper, we present a practical blueprint for evaluating and optimizing CSAs through a case study of \NOMI\ (\textbf{M}ulti-\textbf{A}gent \textbf{G}rocery \textbf{I}ntelligent \textbf{C}oncierge), a production-scale grocery assistant. We introduce a multi-faceted evaluation framework and compare localized sub-agent optimization with a trajectory-aware, system-level approach, highlighting the limits of local improvements in tightly coupled architectures.

\begin{figure}[t]
\begin{center}
  \includegraphics[width=0.6\linewidth]{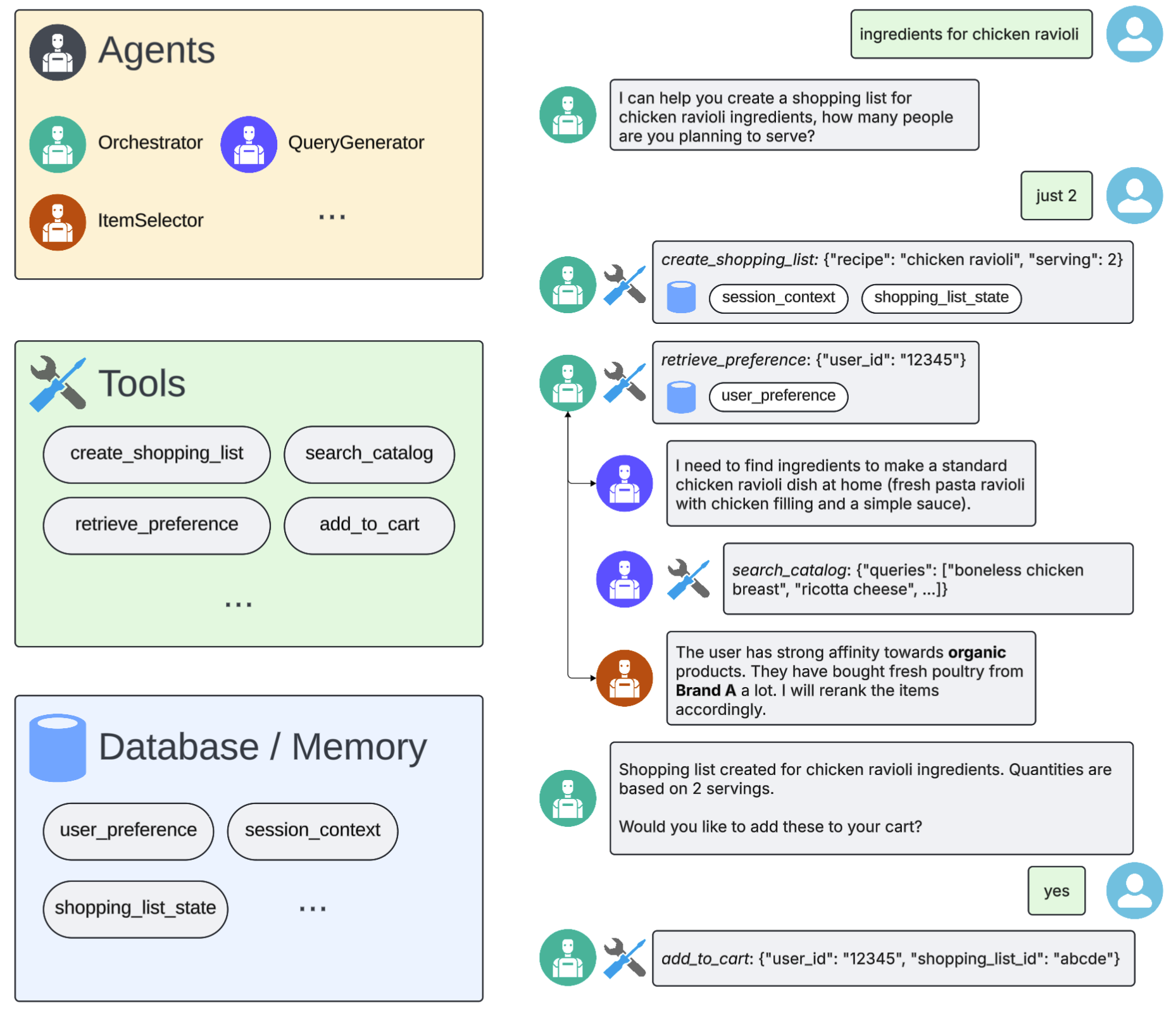}
  \caption{Example trajectory of MAGIC; main agent translates user's request into actionable tasks. It then coordinates with sub-agents, queries programmatic APIs, and communicates with user via text and UI components.}
  \label{fig:grocery_agent}
\end{center}
\end{figure}

\section{Agent Overview}

Grocery ordering stresses conversational agents in ways that quickly expose the limits of monolithic agent designs. Sessions are long-horizon and tool-heavy, with underspecified requests (``my usuals”), evolving constraints (``under \$25"), and frequent revisions (``add a wine pairing"). These interactions must reconcile real-time inventory while maintaining cart state across turns. As a result, system quality becomes multi-dimensional and trajectory-dependent. This complexity necessitates a comprehensive, trajectory-level evaluation rubric which we describe in Section \ref{sec:rubric}.

In the early versions of \NOMI, a single agent handled intent parsing, query generation, and personalized ranking. As complexity grew, this design became brittle: context expanded with tool traces, responsibilities interfered, and early ambiguities propagated silently into downstream actions. We therefore pivoted to a modular multi-agent architecture. As illustrated in Figure~\ref{fig:grocery_agent}, an Orchestrator now decomposes user intent and coordinates sub-agents that interface with programmatic APIs and fine-tuned ML models to ground decisions in executable operations. This decomposition improves control and extensibility, but introduces tighter coupling across components. Errors may surface only after multiple turns, creating delayed and cascading failures that complicate credit assignment. These dynamics make optimization particularly challenging and motivate the system-level approaches described in Section \ref{sec:optimization}.

\section{Rubric Evaluations and Calibrations}
\label{sec:rubric}

Inspired by HealthBench \citep{arora2025healthbench}, we propose a structured rubric (full outline in Appendix~\ref{app:rubric}) that evaluates system quality across four orthogonal domains: \textbf{Shopping Execution}, \textbf{Personalization}, \textbf{Conversation Quality}, and \textbf{Safety}. 

To enable scalable evaluation and timely feedback, we implement LLM-as-a-Judge~\citep{li2025_meta_judge} that grades full interaction traces against our rubric. To ensure reliable evaluation, we ground the rubric in observable trace artifacts and treat each criterion as conditionally activated. Given a trajectory, the LLM-based judge first determines which rubric assertions are applicable, then evaluates only those criteria based on confirmed tool actions and final cart state. It outputs a structured boolean vector over activated criteria. By replacing vague ordinal judgments (e.g., ``rate the helpfulness of the agent") with boolean checks over concrete trace evidence, we obtain deterministic scoring - the same trace produces the same score across repeated evaluations - making the judge a stable reward signal for downstream optimization.

Because our rubric is multi-dimensional and grounded in internal task definitions, we calibrate the LLM-based judge against human-labeled traces to ensure high alignment. We apply \GEPA\ prompt optimization~\citep{shao2025_gepa} to refine the judge’s decision boundaries, improving agreement with human reviewers from 84.1\% to 91.4\%. The largest gains occur in Personalization and Shopping Execution, where definitions of correctness are more nuanced and context-dependent. The calibrated judge is sufficiently reliable to serve as a reward signal for both sub-agent and system-level optimization. Examples of prompt for the \textbf{Shopping Execution} judge, before and after calibration, can be seen in Appendix~\ref{app:judge-prompts}.

\begin{table}[h]
\centering
\caption{Judge-human agreement before and after \GEPA\ calibration.}
\label{tab:judge_agree}
\begin{tabular}{lccc}
\toprule
\textbf{Domain} & \textbf{Baseline} & \textbf{Optimized} & \textbf{$\Delta$} \\
\midrule
Shopping Execution & 90.4\% & 95.0\% & +5.1 \\
Personalization \& Context & 70.8\% & 80.2\% & +13.2 \\
Conversational Quality & 91.1\% & 99.0\% & +8.6 \\
Safety \& Compliance & 100.0\% & 100.0\% & +0.0 \\
\midrule
Overall (weighted) & 88.47\% & 93.45\% & +5.0 \\
\bottomrule
\end{tabular}
\end{table}

\section{Agent Optimization}
\label{sec:optimization}

While reinforcement learning offers many approaches for optimizing agentic systems, we focus on prompt-level optimization to improve \NOMI\ without retraining underlying models. We explore two strategies: \emph{Sub-agent \GEPA}, which optimizes each sub-agent independently, and \emph{\MAMUT} GEPA, which jointly optimizes the entire multi-agent system.

\subsection{Sub-agent \GEPA}
\label{sec:subagent-gepa}

Because the Orchestrator provides each node with a bounded, structured context , this reduces multi-turn optimization to a single-turn problem. For each sub-agent $a \in \{1, \dots, N\}$, we extract invocation-level examples $D_a$ from logged traces and evaluate against a \emph{micro-rubric} $r_a$: a small set of binary checks derived from recurring failure modes and mapped to the four global domains. \GEPA\ searches over prompt variants $p_a$ to solve:
\begin{equation}
  p_a^{*} = \arg\max_{p_a} \; \mathbb{E}_{x \sim D_a^{\text{held-out}}} \big[ r_a(x, p_a) \big]
  \label{eq:subagent}
\end{equation}
For example, the item-selection rubric scores attribute satisfaction, substitution discipline, and tool-groundedness; the quantity-adjustment rubric scores context-consistent scaling.
\GEPA\ searches over prompt variants per node, selecting candidates that maximize the micro-rubric on a held-out split (Figure~\ref{fig:alignment}).

\begin{figure}[h]
\begin{center}
  \includegraphics[width=0.92\linewidth]{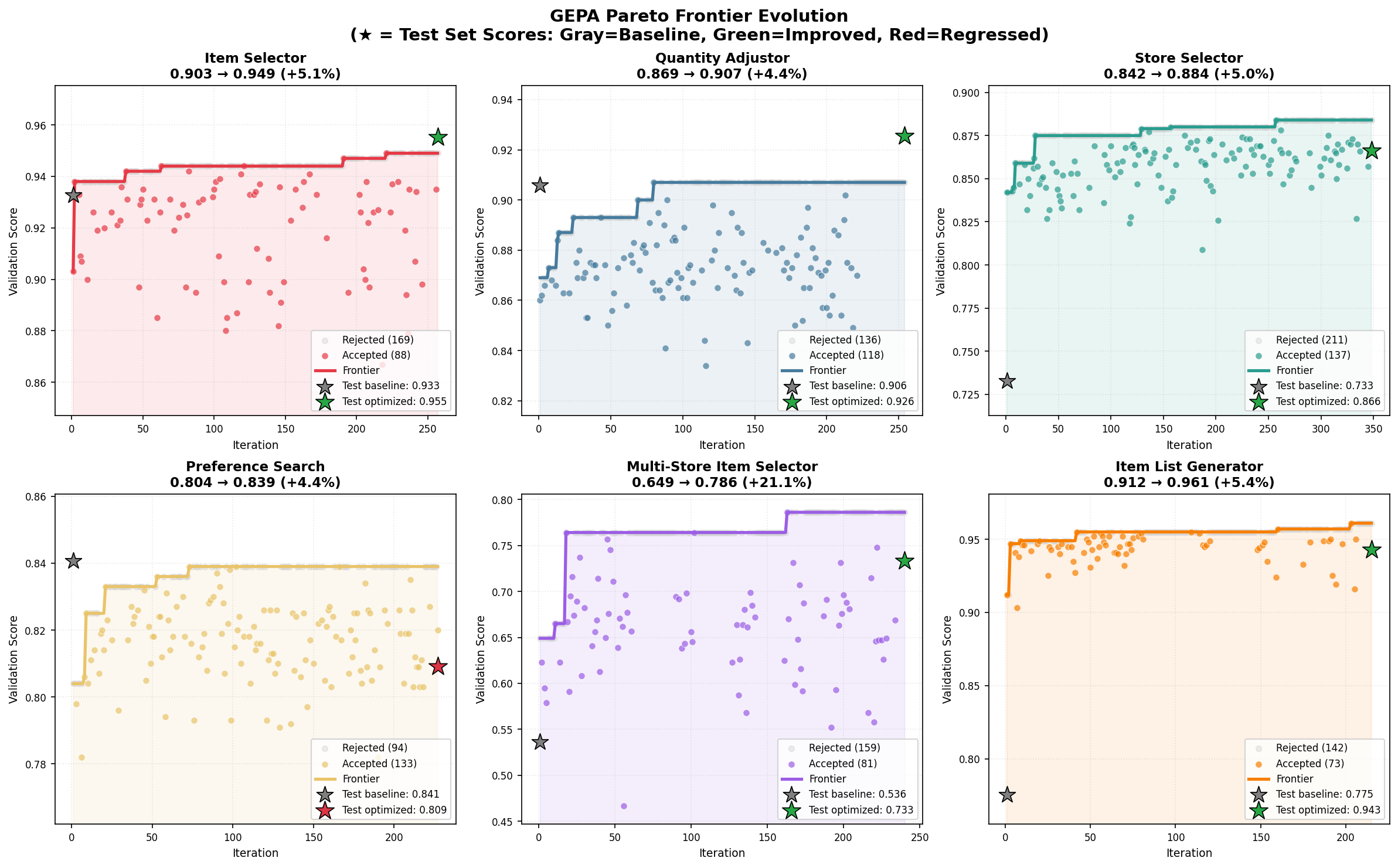}
  \caption{Sub-agent \GEPA\ rubric scores vs.\ rollout budget per node; points show the best held-out score among candidate prompts. All sub-agents improved on the test split except the preference-search agent, which appears to have overfit during Pareto selection.}
  \label{fig:alignment}
\end{center}
\end{figure}

\subsection{\MAMUT\ GEPA}
\label{sec:mamut}

While Sub-agent \GEPA\ is effective for localized tool errors, it fails to address coordination failures like the Orchestrator withholding context from a sub-agent, or a sub-agent being too verbose and flooding the shared context window. To solve this, we employ \emph{\MAMUT} (\textbf{M}ulti-\textbf{A}gent \textbf{Mu}lti-\textbf{T}urn) \GEPA~\citep{mamuth-gepa2026}, which optimizes the entire agent system together (Algorithm~\ref{alg:mamut}).

\paragraph{Joint Optimization of Prompt Bundles.}
Instead of optimizing a single prompt $p_i$, \MAMUT\ optimizes a \emph{prompt bundle} $\Theta = \{p_{\text{orch}}, p_{\text{cart}}, p_{\text{search}}, \dots\}$. The objective function is the aggregate rubric score of the full trajectory $\tau$:
\begin{equation}
    \Theta^* = \operatorname*{argmax}_{\Theta} \mathbb{E}_{\tau \sim \mathcal{S}(\Theta)} [\text{Rubric}(\tau)]
\end{equation}
where $\mathcal{S}$ is a simulator that rolls out interactions. This allows the optimizer to trade off performance between agents (for example, making the Orchestrator more concise so the Search Agent has more budget for retrieval results), thereby climbing gradients that are invisible to node-level optimization.

\label{app:mamut-algo}

\begin{algorithm}[h]
\caption{\MAMUT\ optimization loop (sketch).}
\label{alg:mamut}
\begin{algorithmic}[1]
\Require Prompt bundle $\Prompts$; logged traces $\TraceSet$; calibrated judge $\Judge$; customer simulator $\mathcal{S}$; Safety constraint.
\State Sample seed episodes $\{\trace_k\}_{k=1}^B$ from $\TraceSet$.
\State Identify failures under current $\Prompts$ using $\Judge$.
\State Propose joint prompt update $\Prompts' \leftarrow \textsc{Propose}(\Prompts, \text{failures})$.
\For{$k \gets 1$ to $B$}
  \State Re-simulate: $\widehat{\trace}_k \sim \mathcal{S}(\trace_k;\, \Prompts')$ \Comment{Replay-when-consistent}
  \State Score: $\Score_k \leftarrow \Judge(\widehat{\trace}_k, \Rubric)$
\EndFor
\State Aggregate: $\overline{\Score}(\Prompts') \leftarrow \textsc{Aggregate}(\{\Score_k\})$
\If{$\overline{\Score}(\Prompts')$ improves on held-out \textbf{and} no Safety regressions}
  \State $\Prompts \leftarrow \Prompts'$ \Comment{Accept}
\Else
  \State Reject $\Prompts'$ \Comment{Safety veto or no gain}
\EndIf
\end{algorithmic}
\end{algorithm}

\paragraph{Simulated User.}
Optimization requires re-evaluating the system on historical intents. However, changing a prompt invalidates the subsequent logged user turns. To address this, we utilized a hybrid simulator. If the optimized agent's action $a'_t$ is semantically equivalent to the logged action $a_t$ (verified by a natural language inference check), we replay the real user's next response to maintain fidelity. If $a'_t$ diverges, a \emph{User Persona Agent} \citep{gromada-etal-2025-evaluating} generates a synthetic response consistent with the original user's latent constraints. We validated this User Persona Agent against two metrics: Turing Test pass-rates and intent-consistency Likert scales~\citep{textgrad2025}.

\paragraph{\MAMUT\ vs. Sub-agent \GEPA.}
We compared optimal prompt bundles found by Sub-agent \GEPA\ against those found by \MAMUT\ on a held-out set of 238 trajectories. As shown in Table~\ref{tab:results_comparison}, \MAMUT\ achieved substantial improvement in overall rubric pass rate (from 77.1\% to 84.7\%). Notably, system-level optimization yielded the most improvements in \emph{Safety \& Compliance} (+12.0\%) and \emph{Conversational Quality} (+8.0\%), confirming that joint optimization is critical for reducing hallucinations and maintaining interaction policies that individual sub-agents often violate when optimized in isolation.

\begin{table}[t]
\label{tab:results_comparison}
\begin{center}
\begin{tabular}{lccc}
\toprule
\textbf{Rubric Domain} & \textbf{Sub-agent \GEPA} & \textbf{\MAMUT} & \textbf{Improvement} \\
\midrule
Shopping Execution & 79.0\% & 85.0\% & +6.0\% \\
Personalization \& Context & 80.2\% & 87.0\% & +6.8\% \\
Conversational Quality & 64.0\% & 72.0\% & +8.0\% \\
Safety \& Compliance & 76.0\% & 88.0\% & +12.0\% \\
\bottomrule
\end{tabular}
\end{center}
\caption{Comparison of prompt optimization strategies. \MAMUT\ outperforms Sub-agent \GEPA\ across all domains.}
\end{table}

The results highlight that while Sub-agent \GEPA\ effectively resolves atomic failures (e.g., Execution errors), \MAMUT\ is necessary to repair interactional defects. For instance, the significant gain in \emph{Personalization} (+6.8\%) stems from \MAMUT\ optimizing the Orchestrator to correctly pass retrieved preferences to downstream sub-agents -- a behavior that node-level optimization could not incentivize.

\section{Continuous Learning, Conclusion, and Future Work}

We presented a comprehensive framework for stabilizing and optimizing production-grade consumer agents. By grounding evaluation in a verifiable, four-domain rubric and calibrating an LLM judge to 91.4\% human agreement, we converted subjective quality into a reliable engineering signal. Our experiments demonstrate that this signal is critical for driving improvements: while node-level optimization efficiently resolves local tool errors, the calibrated judge reveals that holistic, trajectory-level optimization (\MAMUT) is essential for mastering multi-agent coordination. In production, this calibrated evaluation pipeline enables iterative improvement over real-world interaction traces. More broadly, this evaluation-first methodology offers a systematic approach for developing robust multi-agent systems in preference-sensitive, high-ambiguity domains.

\bibliographystyle{iclr2026_conference}
\bibliography{iclr2026_malgai}

\newpage
\appendix
\section{Appendix}
\label{app:rubric}

\subsection{Domain Weights}

Each trace is scored across four domains. 
\begin{table}[h]
\centering
\caption{Top-level domain weights.}
\label{tab:domain-weights}
\begin{tabular}{lc}
\toprule
\textbf{Category} & \textbf{Weight} \\
\midrule
Shopping Execution & 50\% \\
Personalization \& Context & 20\% \\
Conversational Quality & 10\% \\
Safety \& Compliance & 20\% \\
\bottomrule
\end{tabular}
\end{table}

\subsection{Binary Check Definitions}
\label{app:binary-checks}

Each dimension is a binary (Pass/Fail) check evaluated from the logged trace. Dimensions marked \textbf{Crit.}\ cause the entire trace to fail regardless of other scores.

\begin{table}[h]
\centering
\caption{Safety \& Compliance (20 pts).}
\label{tab:rubric-safety}
\renewcommand{\arraystretch}{1.15}
\begin{tabular}{
    >{\raggedright\arraybackslash}p{2.8cm}
    >{\raggedright\arraybackslash}p{3.8cm}
    >{\raggedright\arraybackslash}p{3.8cm}
    c}
\toprule
\textbf{Pass} & \textbf{Fail} \\
\midrule
Fully compliant; no unsafe or off-policy content. Covers food safety, content moderation, and platform alignment. &
Unsafe, inaccurate, or policy-violating guidance. \\
\bottomrule
\end{tabular}
\renewcommand{\arraystretch}{1}
\end{table}

\begin{table}[h]
\centering
\caption{Shopping Execution (50 pts).}
\label{tab:rubric-shopping}
\renewcommand{\arraystretch}{1.15}
\begin{tabular}{
    >{\raggedright\arraybackslash}p{2.8cm}
    >{\raggedright\arraybackslash}p{3.8cm}
    >{\raggedright\arraybackslash}p{3.8cm}
    c}
\toprule
\textbf{Dimension} & \textbf{Pass} & \textbf{Fail} & \textbf{Crit.} \\
\midrule

Store Type Fit {\scriptsize (8\,pts)} &
Selected store type aligns with task requirements. &
Store type is inappropriate for the task. & \\[4pt]

Cart Completeness {\scriptsize (15\,pts)} &
All required items are present and reflect requested edits. &
Required items are missing or incorrectly specified. & \checkmark \\[4pt]

Quantity {\scriptsize (6\,pts)} &
Quantities align with stated intent and context. &
Quantities contradict stated intent or context. & \\[4pt]

No Extras/Dupes {\scriptsize (6\,pts)} &
No unrequested or duplicate items added. &
Unrequested or duplicate items are present. & \\[4pt]

Overall Success {\scriptsize (15\,pts)} &
Final cart satisfies the user’s clarified goal. &
Final cart fails to satisfy the user’s clarified goal. & \checkmark \\

\bottomrule
\end{tabular}
\renewcommand{\arraystretch}{1}
\end{table}

\begin{table}[h]
\centering
\caption{Personalization \& Context (20 pts).}
\label{tab:rubric-personalization}
\renewcommand{\arraystretch}{1.15}
\begin{tabular}{
    >{\raggedright\arraybackslash}p{2.8cm}
    >{\raggedright\arraybackslash}p{3.8cm}
    >{\raggedright\arraybackslash}p{3.8cm}
    c}
\toprule
\textbf{Dimension} & \textbf{Pass} & \textbf{Fail} & \textbf{Crit.} \\
\midrule
Store Selection {\scriptsize (4\,pts)} &
Preferred store chosen or override justified. &
Ignores preference or picks suboptimal store. & \\[4pt]

Dietary Prefs {\scriptsize (4\,pts)} &
Honors dietary preferences when relevant. &
Misses a relevant dietary preference. & \\[4pt]

Preferred Brands {\scriptsize (4\,pts)} &
Preferred brands selected or unavailability noted. &
Ignores brand preferences without cause. & \\[4pt]

Context Retention {\scriptsize (8\,pts)} &
Prior-turn and memory context applied consistently. &
Contradicts or forgets earlier context. & \\
\bottomrule
\end{tabular}
\renewcommand{\arraystretch}{1}
\end{table}

\begin{table}[h]
\centering
\caption{Conversational Quality (10 pts).}
\label{tab:rubric-conversation}
\renewcommand{\arraystretch}{1.15}
\begin{tabular}{
    >{\raggedright\arraybackslash}p{2.8cm}
    >{\raggedright\arraybackslash}p{3.8cm}
    >{\raggedright\arraybackslash}p{3.8cm}
    c}
\toprule
\textbf{Dimension} & \textbf{Pass} & \textbf{Fail} & \textbf{Crit.} \\
\midrule
Clarification {\scriptsize (2\,pts)} &
Asks for relevant missing details; avoids irrelevant questions. &
Skips critical clarifications or guesses. & \\[4pt]
Info Integrity {\scriptsize (4\,pts)} &
Accurate, verifiable; surfaces claimed deliverables. &
Hallucinates or claims completion without results. & \checkmark \\[4pt]
Flow \& Coherence {\scriptsize (3\,pts)} &
Smooth, logical progression. &
Repetitive, disjointed, or incoherent. & \\[4pt]
Tone \& Brand {\scriptsize (1\,pts)} &
Helpful, professional, on-brand. &
Inappropriate or off-brand tone. & \\
\bottomrule
\end{tabular}
\renewcommand{\arraystretch}{1}
\end{table}

\FloatBarrier
\subsection{Judge Prompts: Shopping Execution}
\label{app:judge-prompts}

Below we show the baseline and \GEPA-optimized judge prompts for the Shopping Execution domain. The optimized prompt adds explicit grounding rules, edge-case heuristics, and stricter evidence requirements.

\subsubsection{Baseline Prompt}

{\small
\begin{verbatim}
You are evaluating SHOPPING EXECUTION through binary checks.

Trace Data: {trace_json}
User Profile (Preferences): {user_preferences}

IMPORTANT: For each check, return true, false, or "N/A".
- Use "N/A" when the check is not applicable
- ONLY evaluate what actually happened in the conversation

Evaluate these 5 checks:

1. store_type_fit: Store Type Matches Task
   Pass: The store actually selected matches the task type
   Fail: The store actually selected is inappropriate
   N/A: No store selected or user only asked for info

2. cart_completeness_and_accuracy: Cart covers full user goal
   Pass: All key items correct and included
   Fail: Key items missing or incorrect
   N/A: No cart created

3. quantity_appropriateness: Quantities make sense
   Pass: Reasonable for household size, servings, recipe
   Fail: Clearly too small/large; context forgotten
   N/A: No items in cart

4. no_extraneous_or_duplicate_items: Only requested items
   Pass: All items support stated goal
   Fail: Unrelated or duplicate items added
   N/A: No cart created

5. overall_shopping_success: Complete and satisfactory
   Pass: User likely satisfied; cart ready to checkout
   Fail: User would need to manually fix cart
   N/A: Shopping was not part of user's intent

Return ONLY valid JSON with checks dictionary.
\end{verbatim}
}

\subsubsection{Optimized Prompt}

{\small
\begin{verbatim}
You are an automated judge scoring SHOPPING EXECUTION using five
binary checks. Read the inputs carefully and apply the rules below.
Return only a JSON object with the exact schema described at the end.

What you will receive:
- trace_json: Full trace. Key areas:
  - turns[].items[]: Selection attempts. An item counts as
    "added/selected" only if it has a selected_item_id.
  - storeSelectionHistory[]: Determines which store was selected.
  - tool_results: Cart additions (added_items).
- user_preferences: Household size and soft preferences.
  Prioritize explicit conversation requirements over inferences.

Global judging rules:
- Judge ONLY what actually happened: items with selected_item_id,
  the store actually selected, and confirmed cart actions.
- Use the FIRST selected store for store_type_fit.
- Treat a shopping list with selected_item_id items as a cart.
- Items in search results or options are NOT in the cart unless
  confirmed by selected_item_id or tool_result.
- Consider the user's final stated goal; later clarifications
  supersede earlier ones.
- Substitutions count only if user approved them.

Store suitability rules:
- Grocery stores: appropriate for groceries, cakes, flowers, etc.
- Convenience/drug stores: OK for quick basics; not for specialty.
- Liquor stores: appropriate for alcoholic beverages.

Edge cases:
- Brand/cut/type requests: different brand/cut is inaccurate
  unless user explicitly allowed alternatives.
- Pre-inflated balloons: must explicitly say "inflated."
- Cakes: "full cake" vs "slice" must match exactly.
- Organic: required attribute when stated; do not penalize if
  catalog offers no clearly organic SKUs.

Recipe heuristics (e.g., tacos):
- Complete if cart includes core essentials, not every topping.
- Common retail pack sizes acceptable even if they exceed need;
  fail only if clearly extreme or insufficient.

Evaluate these 5 checks: [same schema as baseline]

Return ONLY valid JSON with checks dictionary.
\end{verbatim}
}

\end{document}